\documentclass[sigconf,natbib=true]{acmart}
\AtBeginDocument{%
  }

\copyrightyear{2026}
\acmYear{2026}
\setcopyright{cc}
\setcctype{by}
\acmDOI{10.1145/3805712.3808634}
\acmConference[SIGIR '26]{Proceedings of the 49th International ACM SIGIR Conference on Research and Development in Information Retrieval}{July 20--24, 2026}{Melbourne, VIC, Australia}
\acmBooktitle{Proceedings of the 49th International ACM SIGIR Conference on Research and Development in Information Retrieval (SIGIR '26), July 20--24, 2026, Melbourne, VIC, Australia}
  
\acmISBN{979-8-4007-2599-9/2026/07}

\usepackage[table]{xcolor} 
\usepackage{enumitem}
\usepackage{booktabs}   
\usepackage{multirow}   
\usepackage{graphicx}   
\usepackage{amsmath}
\usepackage{mathtools}
\usepackage{tabularx}
\usepackage{arydshln}
\usepackage{wasysym}
\usepackage{makecell}
\usepackage{array}
\usepackage{threeparttable}
\usepackage{adjustbox}
\usepackage{colortbl}
\usepackage{savesym}
\usepackage{amssymb}
\usepackage{wasysym}
\usepackage{fontawesome5}
\usepackage{setspace}
\usepackage{xcolor}

\newcommand{\ie}{\emph{i.e., }}
\newcommand{\eg}{\emph{e.g., }}

\DeclareRobustCommand{\quoted}[1]{``#1''}

\begin{document}


\title{AlpsBench: An LLM Personalization Benchmark for Real-Dialogue Memorization and Preference Alignment}

\author{Jianfei Xiao$^{1}$, Xiang Yu$^{1}$, Chengbing Wang$^{1}$, Wuqiang Zheng$^{1}$, Xinyu Lin$^{2}$, Kaining Liu$^{1}$, Hongxun Ding$^{1}$, Yang Zhang$^{2}$, Wenjie Wang$^{1}$, Fuli Feng$^{1}$, and Xiangnan He$^{1,*}$}

\affiliation{
\institution{$^1$University of Science and Technology of China, $^2$National University of Singapore. $^*$Corresponding author}
\country{}
}
\email{jianfeixiao@mail.ustc.edu.cn, hexn@ustc.edu.cn}
\renewcommand{\shortauthors}{Xiao et al.}

\begin{abstract}
  As Large Language Models (LLMs) evolve into lifelong AI assistants, 
LLM personalization has become a critical frontier. 
However, progress is currently bottlenecked by the absence of a gold-standard evaluation benchmark. 
Existing benchmarks either overlook personalized information management that is critical for personalization or rely heavily on synthetic dialogues, which exhibit an inherent distribution gap from real-world dialogue. 
To bridge this gap, we introduce \textbf{AlpsBench}, \textbf{A}n \textbf{L}LM \textbf{P}er\textbf{S}onalization benchmark derived from real-world human--LLM dialogues. 
AlpsBench comprises 2,500 long-term interaction sequences curated from WildChat, paired with human-verified structured memories that encapsulate both explicit and implicit personalization signals. 
We define four pivotal tasks---personalized information \textit{extraction}, \textit{updating}, \textit{retrieval}, and \textit{utilization}---and establish protocols to evaluate the entire lifecycle of memory management. Our benchmarking of frontier LLMs and memory-centric systems reveals that: (i) models struggle to reliably extract latent user traits; (ii) memory updating faces a performance ceiling even in the strongest models; (iii) retrieval accuracy declines sharply in the presence of large distractor pools; and (iv) while explicit memory mechanisms improve recall, they do not inherently guarantee more preference-aligned or emotionally resonant responses. AlpsBench aims to provide a comprehensive framework to accelerate research toward truly personalized AI assistants.

\end{abstract}



\maketitle

\section{Introduction}

Recent advancements in LLMs (\eg long-context understanding and self-evolution) have demonstrated promising results across different domains~\cite{wang2025benchmark, voyager, star}, opening up significant potential for LLMs to shift from instant general-purpose tools to lifelong evolving AI assistants. 
Despite that mainstream LLMs are proficient in addressing generic tasks, they still struggle to accommodate heterogeneous users' needs, potentially hurting the user experience in daily use of AI~\cite{salemi2024lamp, personasurvey}. 
In the light of this, enabling LLM personalization becomes crucial to achieve lifelong personal intelligence, strongly attracting and motivating both academia~\cite{pers0,pers1,pers2,pers3} and industry (\eg OpenAI~\cite{openai_power_personalized_ai_2025}, Google~\cite{lee2025personalized}, and Anthropic~\cite{anthropic_assistant_axis_2026}) to explore personalizing LLM responses for different individuals.

\begin{table}[t]
\centering
\caption{Comparison between AlpsBench and existing LLM personalization benchmarks. \quoted{Real} denotes whether the benchmark is built upon real-world data. Columns \quoted{T1} to \quoted{T4} represent whether the benchmarks contain corresponding testing dimensions.}
\vspace{-0.3cm}
\label{tab:benchmark_compare}

\definecolor{colorHist}{HTML}{EAD4D4} 
\definecolor{colorPers}{HTML}{FFF4E5} 
\definecolor{colorDial}{HTML}{D4DCEA} 
\definecolor{colorOurs}{HTML}{FFF9C4} 

\resizebox{\columnwidth}{!}{%
\begin{threeparttable}
\renewcommand{\arraystretch}{1.25} 
\setlength{\tabcolsep}{3.5pt}
\begin{tabular}{>{\raggedright\arraybackslash}p{0.48\columnwidth} c c c c c c c c c}
\toprule
\textbf{Benchmark} &
\textbf{Real} &
\textbf{T1} & \textbf{T2} & \textbf{T3} &
\multicolumn{5}{c}{\textbf{T4: Utilization Dimensions}} \\
\cmidrule(lr){6-10}
& & \textbf{Ext.} & \textbf{Upd.} & \textbf{Retr.} & \textbf{PA} & \textbf{PF} & \textbf{VRA} & \textbf{CF} & \textbf{EI} \\
\midrule

\rowcolor{colorHist}
LaMP            & \CIRCLE & \Circle & \Circle & \Circle & \LEFTcircle & \LEFTcircle & \Circle & \Circle & \Circle \\
\rowcolor{colorHist}
PersonalLLM     & \Circle & \Circle & \Circle & \Circle & \LEFTcircle & \CIRCLE & \Circle & \Circle & \Circle \\
\rowcolor{colorHist}
EQ-Bench        & \Circle & \Circle & \Circle & \Circle & \Circle & \Circle & \Circle & \Circle & \CIRCLE \\
\rowcolor{colorHist}
PersoBench      & \Circle & \Circle & \Circle & \Circle & \CIRCLE & \CIRCLE & \Circle & \CIRCLE & \Circle \\
\rowcolor{colorHist}
PersonaFeedback & \Circle & \Circle & \Circle & \Circle & \CIRCLE & \CIRCLE & \Circle & \CIRCLE & \Circle \\
\midrule 

\rowcolor{colorDial}
LoCoMo          & \Circle & \LEFTcircle & \Circle & \CIRCLE & \LEFTcircle & \Circle & \Circle & \Circle & \Circle \\
\rowcolor{colorDial}
LongMemEval     & \Circle & \CIRCLE & \CIRCLE & \CIRCLE & \LEFTcircle & \Circle & \Circle & \Circle & \Circle \\
\rowcolor{colorDial}
PersonaLens     & \Circle & \Circle & \Circle & \LEFTcircle & \CIRCLE & \CIRCLE & \Circle & \CIRCLE & \Circle \\
\rowcolor{colorDial}
HaluMem         & \Circle & \CIRCLE & \CIRCLE & \LEFTcircle & \LEFTcircle & \Circle & \Circle & \Circle & \Circle \\
\rowcolor{colorDial}
PersonaMem v2   & \Circle & \CIRCLE & \CIRCLE & \LEFTcircle & \CIRCLE & \CIRCLE & \Circle & \Circle & \Circle \\

\midrule 

\rowcolor{colorOurs}
\textbf{AlpsBench} & \CIRCLE & \CIRCLE & \CIRCLE & \CIRCLE & \CIRCLE & \CIRCLE & \CIRCLE & \CIRCLE & \CIRCLE \\
\bottomrule
\end{tabular}

\begin{tablenotes}[flushleft]\scriptsize
\item[] \textbf{Category:} 
\textcolor{colorHist}{\rule{7pt}{7pt}} Memory-free Preference Alignment, 
\textcolor{colorDial}{\rule{7pt}{7pt}} Memory-aware Preference Alignment,
\textcolor{colorOurs}{\rule{7pt}{7pt}} Ours.\\
\CIRCLE: Fully Supported / Real Data, \LEFTcircle: Partially Supported, \Circle: Not Supported / Synthetic Data.\\
\textbf{Ext.} = Memory Extraction, \textbf{Upd.} = Memory Updating, \textbf{Retr.} = Memory Retrieval, \textbf{PA} = Persona Awareness, \textbf{PF} = Preference Following, \textbf{VRA} = Virtual-Reality Awareness. \textbf{CF} = Constraint Following. \textbf{EI} = Emotional Intelligence.
\end{tablenotes}
\end{threeparttable}
}
\vspace{-0.5cm}
\end{table}

\begin{figure}[t]
\vspace{-0.2cm}
\setlength{\abovecaptionskip}{-0.2cm}
\setlength{\belowcaptionskip}{-0.3cm}
\centering
\includegraphics[width=\linewidth]{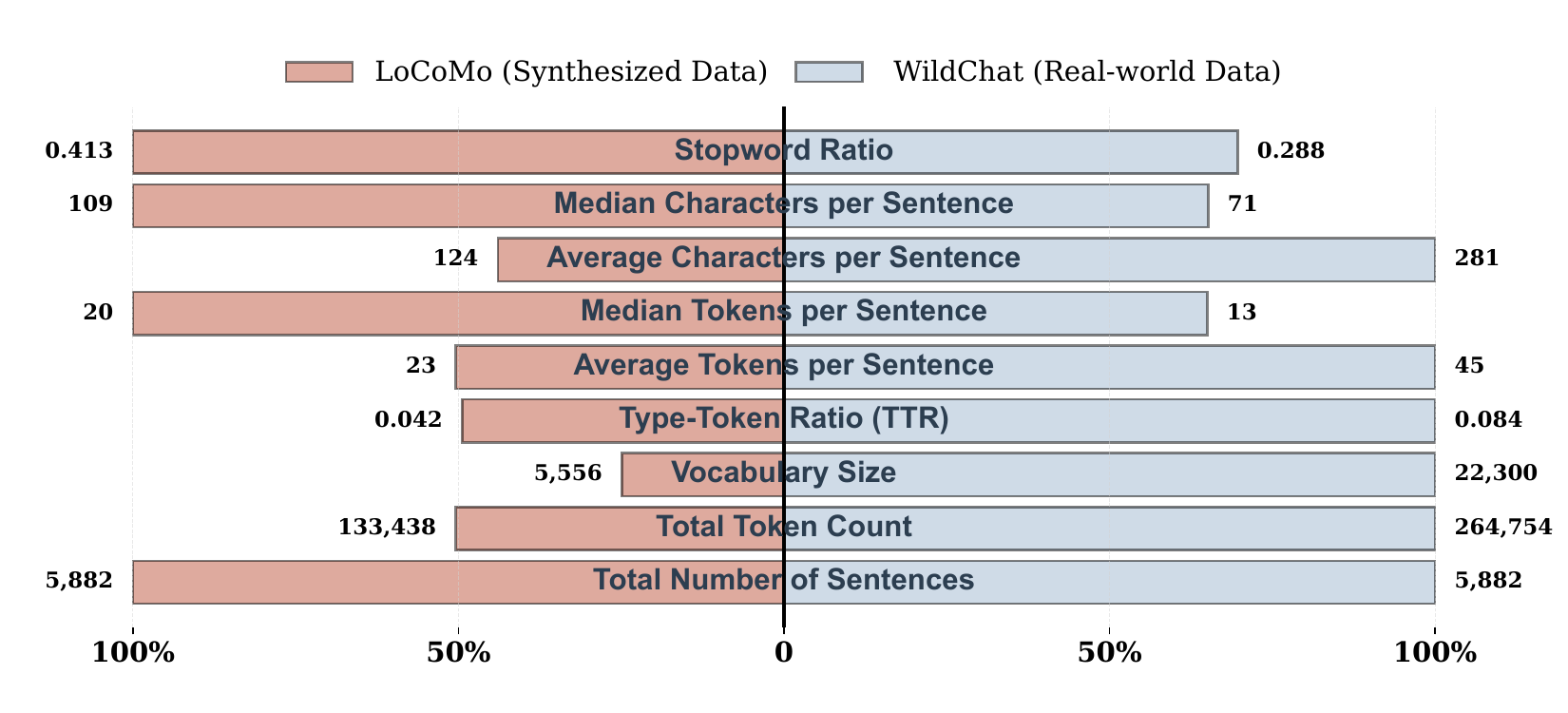}
\caption{Comparison between synthesized and real data.}
\Description{Examples contrasting synthesized and real dialogue data, emphasizing that real data contains more diverse and implicit personalization signals.}
\label{fig:sim_vs_real_sentence}
\end{figure}

\begin{figure}[t]
\vspace{-0cm}
\setlength{\abovecaptionskip}{0cm}
\setlength{\belowcaptionskip}{-0.3cm}
  \centering
\includegraphics[width=0.9\linewidth]{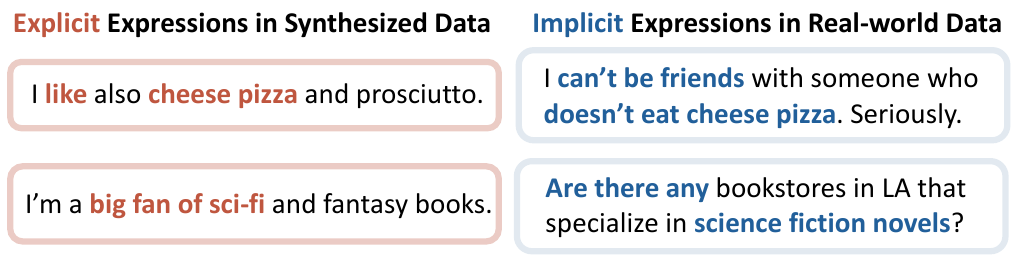}
  \caption{Examples of the synthesized and the real-world human expressions.}
  \Description{Side-by-side examples showing that synthesized expressions tend to be explicit, while real-world human expressions are more indirect and context-dependent.}
  \label{fig:explicit_vs_implicit}
\end{figure}

Facilitating LLM personalization techniques crucially relies on high-quality evaluation benchmarks. 
Existing benchmarks can be broadly categorized into two groups as shown in Table~\ref{tab:benchmark_compare}:
\begin{itemize}[leftmargin=*]
    \item \textit{\textbf{Memory-free}} benchmarks (\eg LaMP~\cite{salemi2024lamp}, PersonalLLM~\cite{zollopersonalllm}) focus on the alignment between LLMs' final output and user preference in downstream tasks, such as personalized email generation. 
    Despite the effectiveness in assessing personalization in response, they overlook the process of personalized information governance (\eg memory extraction, update, retrieval), which is a key component in understanding and leveraging user preference in personalized responses. 
    \item \textit{\textbf{Memory-aware}} benchmarks explicitly incorporate memory evaluation based on synthetic human-LLM dialogues along multiple dimensions (\eg long-term history memorization~\cite{locomo,longmemeval} and emotional preference alignment~\cite{emobench,eqbench}). 
    Nonetheless, they suffer from a critical distributional gap between simulated and real-world data, causing two limitations. 
    1) Lack of diverse conversations, where LLM-synthesized dialogues tend to be homogeneous~\cite{jiang2025artificial}, failing to capture the natural conversational diversity (see Figure~\ref{fig:sim_vs_real_sentence}). 
    2) Lack of implicit expressions. In real-world interactions, users often convey personalized information implicitly. However, synthetic dialogues can be overly explicit and simplistic (see Figure~\ref{fig:explicit_vs_implicit}), failing to generalize to real-world scenarios. 
\end{itemize}

In this light, it is imperative to build benchmarks based on real-world user conversational data that capture both conversational diversity and implicit personalization signals. 
Moreover, instead of simply verifying LLMs' final output, it is essential to systematically evaluate the process of understanding, storing, and utilizing personalized information. 
Hence, we propose a comprehensive evaluation framework with four core tasks. 
\begin{itemize}[leftmargin=*]
    \item \textit{Task 1: Personalized information \textbf{extraction}}, which evaluates the ability to extract personalized information from dialogues. Given the raw conversational data, the evaluated LLMs are asked to distill them into structured memory containing personalized information (refer to Section~\ref{method:task1} for details). In the paper, we report semantic-matching F1 with LLM-as-a-Judge against ground-truth memory, while additional diagnostic metrics are available on the leaderboard. 
    
    \item \textit{Task 2: Personalized information \textbf{update}}, which assesses the capacity to track the dynamics of user information and update structured memory accurately. Given both a historical and a new dialogue, the evaluated LLMs are asked to output updated memory and the memory manipulation (\ie retention, addition, and modification). In the paper, we report semantic-matching F1 for updated memories, while action-level accuracy and other diagnostic metrics are available on the leaderboard. 
    
    \item \textit{Task 3: Personalized information \textbf{retrieval}}, which measures the ability to retrieve relevant personalized information. Given a user query and a candidate set of memories, the evaluated LLMs are asked to select the relevant memory for generating a personalized response. Recall is adopted as the evaluation metric. 
    \item \textit{Task 4: Personalized information \textbf{utilization}}, which examines the alignment of user preference in LLMs' response. We break down the task into measurement of five LLM capabilities, including persona awareness, preference following, virtual-reality awareness, constraint following, and emotional intelligence for a comprehensive alignment of dimensions (refer to Section~\ref{method:task4} for detailed task designs and evaluation metrics). 
\end{itemize}

To construct a real-world comprehensive benchmark, we develop a four-step pipeline. 
\textit{Step 1}: Data collection. We leverage the real-world WildChat dataset~\cite{wildchat} and collect long-term dialogues\footnote{WildChat is distributed under the Open Data Commons Attribution License (ODC-BY). We acknowledge WildChat as the source dataset and make our reprocessed derivative data available on Hugging Face under ODC-BY.}. 
\textit{Step 2}: Memory extraction and filtering. 
We design a structured memory taxonomy to extract personalized information from dialogues and filter redundant or low-signal samples. 
\textit{Step 3}: Task construction. 
Based on the collected dialogues and extracted memories, we construct evaluation data for four tasks, including evaluation queries and corresponding ground truth. 
\textit{Step 4}: Human verification and quality control. 
Human annotators then verify the constructed memories, task instances, and LLM-as-a-Judge alignment to ensure benchmark reliability. 
This pipeline can be continuously updated with evolving real-world human-LLM conversations, enabling a dynamic benchmark. 

Building upon the above pipeline, we introduce \textbf{AlpsBench}, 
a comprehensive benchmark for evaluating LLM personalization, with the following key features: 
1) AlpsBench consists of 2,500 high-quality real-world human–LLM dialogues with 6 to 249 turns. 
2) It covers four tasks for personalized information, \ie extraction, update, retrieval, and utilization. 
3) It assesses multidimensional capabilities of memory utilization, \ie persona awareness, preference following, virtual-reality awareness, constraint following, and emotional intelligence. 
We also release a public leaderboard\footnote{\url{https://misshsiaoo.github.io/Alps_Bench/}.} to facilitate continuous benchmarking from the community.

We benchmark frontier LLMs, including general-purpose and memory-oriented LLMs, and derive several key findings:
(i) models exhibit limited reliability in extracting latent user traits;
(ii) memory updating reaches a performance plateau even for the strongest models;
(iii) retrieval accuracy degrades substantially in the presence of large distractor sets; and
(iv) explicit memory mechanisms do not inherently ensure stronger preference alignment or emotional resonance in responses.
We release our code at \url{https://github.com/ThisIsCosine/AlpsBench}. 


The key contributions of this work are summarized as follows: 
\begin{itemize}[leftmargin=*]
    \item We introduce AlpsBench, a dynamic and extensible LLM personalization benchmark built on real-world human–LLM conversations with human-verified structured memory. 

    \item We propose a systematic evaluation framework comprising four core tasks (\ie personalized information extraction, update, retrieval, and utilization) to benchmark personalization ability. 

    \item Extensive evaluations on AlpsBench reveal that existing models remain limited in reliable long-term personalization and context-aware utilization, highlighting substantial opportunities for future research in personalized LLMs. 
\end{itemize}





\section{AlpsBench Evaluation Framework}
\label{sec:eval_frame}
\begin{figure}[t]
\vspace{-0.2cm}
\setlength{\abovecaptionskip}{-0cm}
\setlength{\belowcaptionskip}{-0.3cm}
  \centering
\makebox[\linewidth][c]{\scalebox{1}[0.833]{\includegraphics[width=1.06\linewidth]{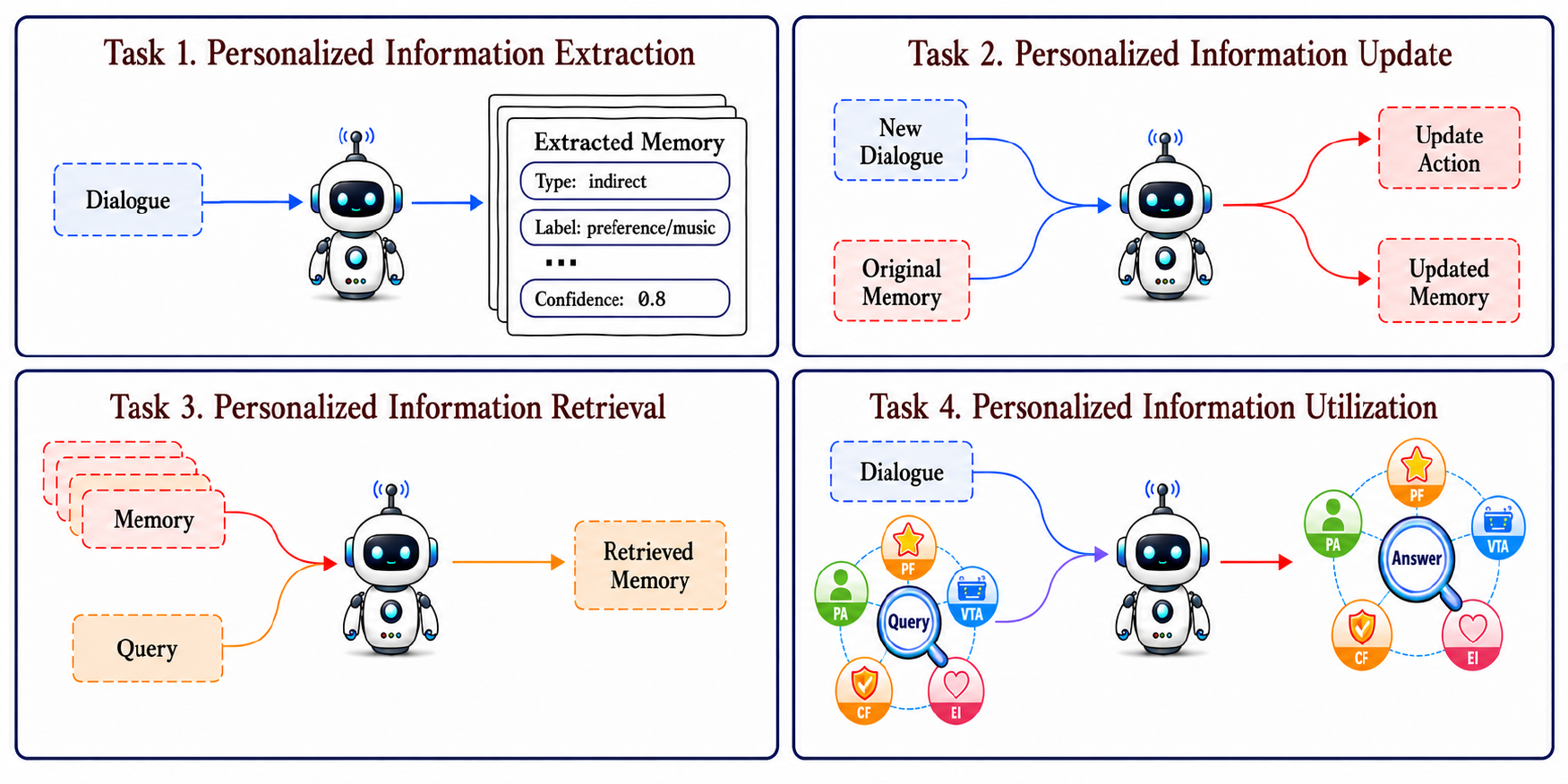}}}
  \caption{The evaluation tasks of AlpsBench. }
  \Description{Overview of the four AlpsBench evaluation tasks, including personalized information extraction, update, retrieval, and utilization.}
  \label{fig:framework}
\end{figure}

To rigorously evaluate the personalization capabilities of AI assistants, we introduce AlpsBench, a comprehensive benchmark designed to simulate real-world personalization scenarios. 
In this section, we provide a detailed description of the task design for AlpsBench (\textit{cf.} Figure~\ref{fig:framework}), which comprises four core tasks specifically designed to assess different facets of personalized AI assistants. 


\subsection{Task 1: Personalized Information Extraction}
\label{method:task1}
A core capability of personalized AI assistants lies in their proficiency in transforming raw conversational data into structured, high-quality memories. This capability reflects the assistants' depth of understanding regarding historical user preferences and key information. Motivated by the need to assess this capability, we propose a \textit{Personalized Information Extraction} task: 
\begin{equation}\small
\quad \mathcal{H} \xrightarrow{\text{AI assistant}} [\hat{\mathcal{M}}_1, \hat{\mathcal{M}}_2, \ldots] ,
\end{equation}
where the AI assistant is tasked with extracting structured user information from the original user dialogue history ($\mathcal{H}$). The resulting output consists of a set of memories ($\hat{\mathcal{M}}_1, \hat{\mathcal{M}}_2, \ldots$), each containing the following attributes:

\begin{itemize}[leftmargin=*]
    \item \textbf{Memory ID:} A unique identifier for the memory entry.
    \item \textbf{Memory Type:} Explicit or implicit preference.
    \item \textbf{Label:} The taxonomic category of the memory.
    \item \textbf{Value:} The exact statement extracted from the original dialogue.
    \item \textbf{Confidence:} The confidence level of the extraction.
\end{itemize}


%
%
%


The quality of the extractions (\ie memories) is evaluated by comparing them against human-annotated ground truth ($\mathcal{M}$). Specifically, we use an LLM-as-a-Judge protocol to semantically match predicted and human-annotated memories one-to-one, computing F1 to capture nuances beyond lexical similarity.





\subsection{Task 2: Personalized Information Update}
\label{method:task2}
The dynamic evolution of user preferences requires personalized AI assistants to possess a sophisticated memory-updating mechanism. This process hinges on three fundamental capabilities: (i) the robustness in filtering dialogue noise, (ii) the ability to add new user preferences, and (iii) the logic for resolving preference conflicts by reasonably updating outdated memories with new information.
Thus, we introduce the \textit{Personalized Information Update} task: 
\begin{equation}\small
\quad (\mathcal{M}, \mathcal{H}_{new}) \xrightarrow{\text{AI assistant}} (\hat{\mathcal{M}}_{new}, \hat{\mathrm{Act}}),
\end{equation}
where the task requires the assistant to output updated memories $\hat{\mathcal{M}}_{new}$ and categorize each update according to an action type $\hat{\mathrm{Act}} \in \{\text{Retention}, \text{Addition}, \text{Modification}\}$, corresponding to the aforementioned capabilities. In this paper, we apply the same semantic matching protocol as in Task 1 and report F1 between updated memories and human-annotated ground truth as the primary metric. Action-level accuracy and other diagnostic metrics are available on the leaderboard. The updated memories can also be leveraged for further analysis of the AI assistants (\textit{cf.} Subsection~\ref{sec:Task2_exp_result}).

\subsection{Task 3: Personalized Information Retrieval}
\label{method:task3}
The ability to retrieve relevant memories is a cornerstone of an AI assistant's personalization. Accurately identifying useful insights from a user's vast memory pool significantly enhances the AI assistant's capacity to provide tailored responses. 
In this light, we introduce the \textit{Personalized Information Retrieval} task: 
\begin{equation}\small
\quad (\mathcal{M}_{\mathrm{pos}}, \{\mathcal{M}_{\mathrm{neg},i}\}_{i=1}^{K}, \mathcal{Q}) \xrightarrow{\text{AI assistant}} \hat{\mathcal{M}}_{\mathrm{pos}},
\end{equation}
where the AI assistant is expected to retrieve the most relevant memory ($\hat{\mathcal{M}}_{\mathrm{pos}}$) based on the user query ($\mathcal{Q}$) from a set of candidate memories, which consists of one positive sample ($\mathcal{M}_{\mathrm{pos}}$) labeled by humans and $K$ randomly sampled negative samples $\{\mathcal{M}_{\mathrm{neg},i}\}_{i=1}^{K}$. We use the widely recognized metric of recall to report the evaluation results of this task. 

\subsection{Task 4: Personalized Information Utilization}
\label{method:task4}
The ability to utilize users’ personal information to address real-world needs is fundamental to personalized AI assistants. To systematically evaluate this capability, we introduce the \textit{Personalized Information Utilization} task, formulated as follows:
\begin{equation}\small
\quad (\mathcal{H}, \mathcal{Q}) \xrightarrow{\text{AI assistant}} \hat{\mathcal{R}} ,
\end{equation}
where the assistant is tasked to produce a response $\hat{\mathcal{R}}$ to a user query $\mathcal{Q}$ based on dialogue history $\mathcal{H}$.

Based on the characteristics of users’ real-world needs, we design five core evaluation dimensions for this task:

\noindent\textbf{Persona Awareness (PA).} This dimension assesses whether the assistant correctly recalls and applies explicit user attributes, such as educational background. Users often provide explicit persona information across multiple sessions, requiring the AI assistant to seamlessly integrate this data into its responses. 


\noindent\textbf{Preference Following (PF).} This dimension measures whether the assistant can capture implicit user preferences within dialogue history and align its responses accordingly. Beyond explicit persona information recall, it tests inductive reasoning to align responses with potential tastes and habits—a critical capability for personalized recommendation and planning. 






\noindent\textbf{Virtual-Reality Awareness (VRA).} This dimension evaluates whether the assistant can distinguish real user information from role-play or fictional content, ensuring that in-character data does not contaminate real-world assistance. For example, if a middle-aged professional previously role-played as a student, the assistant should disregard student's attributes when drafting his/her resume.


\noindent\textbf{Constraint Following (CF).} 
This dimension examines whether the assistant respects constraints expressed by the user in prior interactions. There may be instances where the user explicitly requests that certain information be excluded or specific conditions be met. The assistant should consistently follow these guidelines to ensure alignment with the user's expectations. 


\noindent\textbf{Emotional Intelligence (EI).} This dimension assesses whether the assistant uses user dialogue history to provide emotionally appropriate responses. An effective AI assistant enables differentiated emotional responses, such as offering encouragement to resilient users and reassurance to more sensitive users. 

In this task, we adopt LLM-as-a-Judge to evaluate AI assistants. For each dimension (except for Emotional Intelligence), the judge determines whether the AI assistant's response meets the specified criteria and assigns a binary score (0 or 1) accordingly. Since Emotional Intelligence requires a more nuanced evaluation, we follow prior evaluation practice~\cite{wang2026perm}, which uses the LLM as a reward model to assign a score on a 1–5 scale.


\begin{figure}[t]
\vspace{-0.2cm}
\setlength{\abovecaptionskip}{-0cm}
\setlength{\belowcaptionskip}{-0.3cm}
  \centering
    \makebox[\linewidth][c]{\scalebox{1}[0.9]{\includegraphics[width=1.08\linewidth]{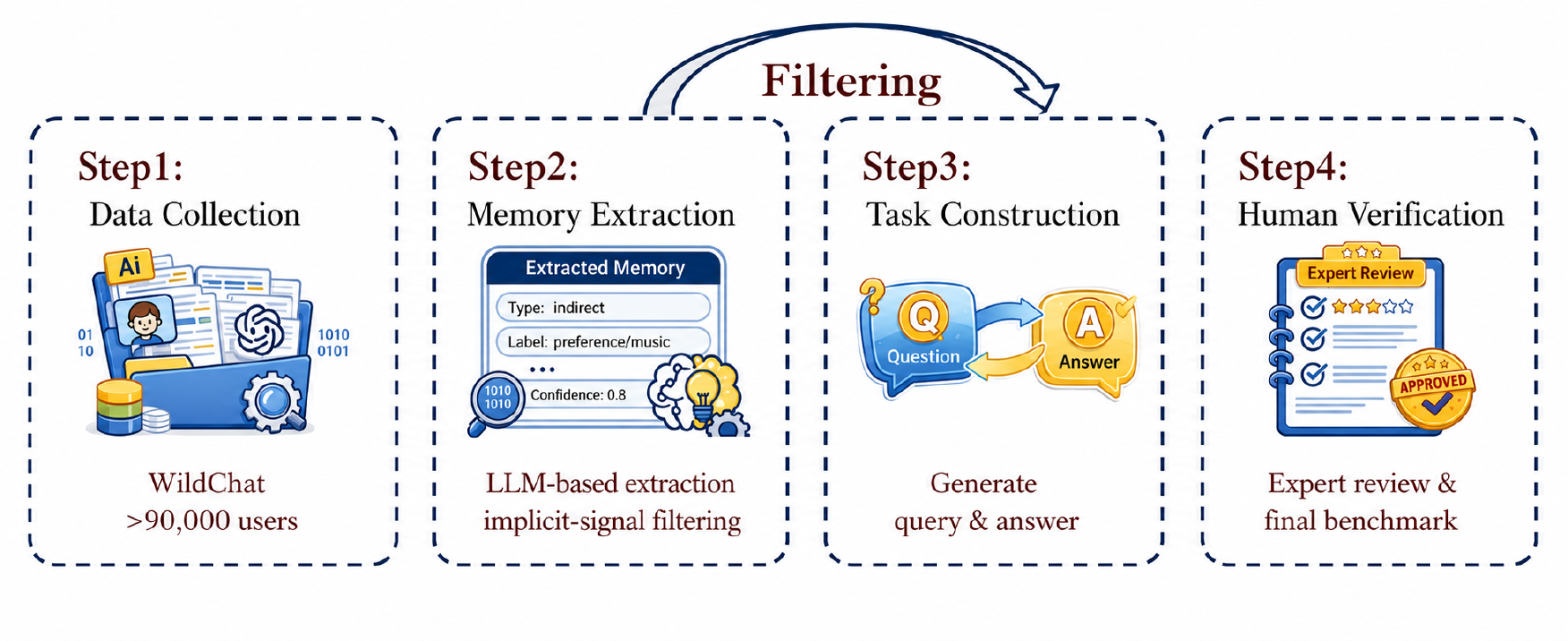}}}
  \caption{The four-step benchmark construction pipeline based on real human–LLM dialogues.}
  \Description{Four-step benchmark construction pipeline: data collection, memory extraction and filtering, task construction, and human verification and quality control.}
  \label{fig:pipeline}
\end{figure}

\section{Curation Pipeline for AlpsBench}
In this section, we describe the four-step curation pipeline of AlpsBench, as depicted in Figure~\ref{fig:pipeline}.


\vspace{1pt}
\noindent$\bullet\quad${\textbf{Step 1: Data Collection}}.
Personalized AI assistants are intended to serve actual human needs. Therefore, to effectively test their capabilities, the design of the personalization benchmark should be grounded in the real-world human-AI interactions. To achieve this, we use the  WildChat~\cite{wildchat}  dataset, which contains approximately 90,000 users’ interaction records covering a broad range of authentic conversational scenarios. Considering longer dialogues inherently carry richer contextual dependencies and cross-temporal information, we further refined the data by  
selecting 
dialogues with more interaction turns.
Each dialogue in this subset contains between 6 and 249 turns of interaction.



\vspace{1pt}
\noindent$\bullet\quad${\textbf{Step 2: Memory Extraction and Filtering}}.
To enable a systematic evaluation of an AI assistant’s ability to process and utilize personalized information, we first employ the \textit{DeepSeek-v3.2 reasoning model} \cite{deepseek-v3.2} to automatically extract structured information from raw user dialogue histories and format it into memory representations, as illustrated in Section~\ref{method:task1}.


Considering the substantial computational cost of running AI assistants and the expense of high-quality human verification (\textit{cf.} Step 4), we further perform data filtering based on high-level semantic categories in the structured memories (\ie the \textit{label} attribute). Specifically, we subsample instances within each category, ensuring that the number of samples per category does not exceed a predefined upper bound. This strategy effectively removes large amounts of semantically redundant content. 

Moreover, to enhance the challenge and discriminative power of the benchmark, we prioritize retaining users whose memories contain implicit information while performing category-based filtering. Specifically, we retain a user $u$ if there exists at least one memory $\mathcal{M}$ with the type attribute labeled as \textit{implicit}. Formally:
\begin{equation}\small
U_{\text{retained}} = \left\{ u \in U \mid \exists \mathcal{M} \in \mathcal{M}_u : \text{type}(\mathcal{M}) = implicit \right\},
\end{equation}
where $U$ is the set of all users. $\mathcal{M}_u$ is the set of memories associated with user $u$. Ultimately, this process yields 2,500 user samples that constitute the final benchmark.





\vspace{1pt}
\noindent$\bullet\quad${\textbf{Step 3: Task Construction}}.
Based on the collected dialogue histories, we construct evaluation queries and corresponding ground truth for the four tasks described in Section~\ref{sec:eval_frame}. 
Selecting a memory as a target memory ($\mathcal{M}_{tgt}$) and using the original user dialogue ($\mathcal{H}$) as context, we employ \textit{GPT-5.2} \cite{gpt-5.2} to synthesize task-specific queries and ground-truth answers. This process can be formulated as:
\begin{equation}\small
(\mathcal{H}, \mathcal{M}_{tgt}, [*]) \xrightarrow{\text{GPT-5.2}} (\hat{\mathcal{Q}}, \hat{\mathcal{A}}) \xrightarrow{\text{Human}} (\mathcal{Q}, \mathcal{A})
\end{equation}
where $[*]$ represents task-specific auxiliary inputs. In the case of \textit{\textbf{Task 2}}, its $[*]$ represents target update strategies—specifically Retention, Addition, and Modification—to guide the generation of update-oriented queries. For \textit{\textbf{Task 3}}, its $[*]$ represents randomly selected memories as distractors (negative samples) to enhance the realism and complexity of the retrieval tasks.

The output of this process consists of two main elements: $\mathcal{Q}$ and $\mathcal{A}$. Specifically, for \textit{\textbf{Task 1}}, $\hat{\mathcal{Q}}$ consists of system instructions for memory extraction, while $\hat{\mathcal{A}}$ represents the $\mathcal{M}_{tgt}$ itself. For \textit{\textbf{Task 2}}, $\hat{\mathcal{Q}}$ corresponds to the user’s new dialogue history along with the system’s instruction (\textit{e.g.}, \quoted{Please update memory}) and $\hat{\mathcal{A}}$ reflects the applied update strategy. For \textit{\textbf{Task 3}}, $\hat{\mathcal{Q}}$ is a practical user inquiry related to the $\mathcal{M}_{tgt}$, with $\hat{\mathcal{A}}$ as the $\mathcal{M}_{tgt}$. Finally, for \textit{\textbf{Task 4}}, $\hat{\mathcal{Q}}$ is a user query related to the $\mathcal{M}_{tgt}$ topic, and $\hat{\mathcal{A}}$ is the generated response addressing that query.

\noindent$\bullet\quad${\textbf{Step 4: Human Verification and Quality Control}}.
We verify AlpsBench from two aspects: whether the constructed benchmark data provide reliable ground truth, and whether the LLM-as-a-Judge evaluator is aligned with human judgments.

\noindent\textbf{Task Construction Quality.}
We first extracted candidate memories from user dialogues and asked annotators to verify the gold memory entries for Task 1. These verified memories then served as the basis for constructing the remaining benchmark instances, including update-oriented dialogues and actions for Task 2 and query-answer pairs for Tasks 3 and 4, yielding 2,500 verified instances for each of the four tasks. All benchmark instances were independently examined by two experienced annotators under a double-blind setting. The two annotators showed strong inter-annotator agreement: for Tasks 1 and 2, inter-annotator memory-set F1 scores reached 0.763 and 0.844, confidence Spearman correlations were 0.868 and 0.735, and Cohen's $\kappa$ values for memory type, label, and time scope ranged from 0.734 to 0.927. For Tasks 3 and 4, approximately 92\% of instances were deemed high-quality. For instances where annotators disagreed, an additional round of manual adjudication and correction was performed. These adjudicated instances were then used to form the final benchmark dataset.



\begin{figure}[t]
    \centering
    \includegraphics[width=\linewidth]{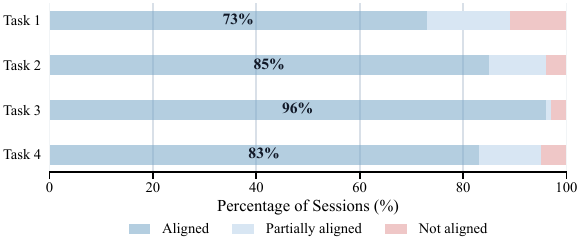}
    {\scriptsize\parbox{0.96\linewidth}{Fully aligned: LLM decision consistent with both annotators. Partially aligned: consistent with one annotator. Not aligned: inconsistent with both.}}
    \caption{Alignment between the LLM-based judge and human experts across four tasks.}
    \Description{Comparison of full, partial, and non-alignment rates between the LLM-based judge and human experts for the four AlpsBench tasks.}
    \label{fig:llm_alignment}
\end{figure}

\noindent\textbf{LLM-as-a-Judge Alignment.}
To support scalable evaluation, we further compare the LLM-based judge with human expert judgments. As shown in Figure~\ref{fig:llm_alignment}, the judge is highly consistent with human preferences across all four tasks, with alignment rates of $96\%$ for Task 3, $85\%$ for Task 2, $83\%$ for Task 4, and $73\%$ for Task 1. These results indicate that the automated evaluation reflects human-centric assessment while remaining scalable to the full benchmark.

\section{Experiments}
\noindent$\bullet\quad${\textbf{Evaluated Models}}.  
For \textbf{general-purpose LLMs}, we include \textbf{open-source models} (DeepSeek-v3.2 in thinking mode \cite{deepseek-v3.2}, Qwen3-max \cite{qwen3}, Llama-4 Maverick \cite{meta_llama4_2025}) and \textbf{closed-source models} (GPT-5.2 \cite{gpt-5.2}, GPT-4.1-mini \cite{gpt4.1}, Gemini-3-Flash-preview \cite{gemini3flash}, Claude-Sonnet-4.5 \cite{anthropic_claude_sonnet45_2025}). For \textbf{memory-oriented systems}, we implemented MemoryOS \cite{kang-2025-memory}, EverMemOS \cite{hu2026evermemos}, A-Mem \cite{xu2025mem}, LightMem \cite{fang2025lightmem}, MemOS \cite{li2025memos}, Mem0 \cite{chhikara2025mem0}, and Mem0g \cite{chhikara2025mem0}. All memory systems use \textbf{GPT-4.1-mini} as the backbone LLM.
For Task 3 in the general-model setting, we additionally evaluate two standard retrieval baselines for memory models, \textbf{nltk+BM25} \cite{nltk,bm25} and \textbf{all-MiniLM-L6-v2} \cite{all-MiniLM-L6-v2}. 
These classic retrieval methods are widely used as external retrievers in memory systems, making them suitable baselines for Task 3.  
\vspace{2pt}
\noindent$\bullet\quad${\textbf{Implementation Details}}.  
We use {GPT-5.2} \cite{gpt-5.2} as the unified {generator} for query generation, and {DeepSeek-v3.2} \cite{deepseek-v3.2} as the {judge model} for automatic evaluation.

\begin{table*}[t]
\centering
\setlength{\abovecaptionskip}{0.0cm}
\setlength{\belowcaptionskip}{0.2cm}
\caption{Tasks 1–4. Experimental evaluation results of general-purpose LLMs.}
\label{tab:evaluation-results}

\setlength{\tabcolsep}{3.5mm}
\resizebox{\textwidth}{!}{%
\begin{tabular}{lcccccccccccccc}
\toprule
\multirow{3}{*}{\textbf{Model}} & \multirow{3}{*}{\textbf{\begin{tabular}[c]{@{}c@{}}Task 1\\ Extraction \end{tabular}}} & \multirow{3}{*}{\textbf{\begin{tabular}[c]{@{}c@{}}Task 2\\ Update \end{tabular}}} & \multicolumn{5}{c}{\textbf{Task 3 Retrieval}} & \multicolumn{7}{c}{\textbf{Task 4 Utilization}} \\ \cmidrule(lr){4-8} \cmidrule(lr){9-15} 
 &  &  & \multicolumn{5}{c}{Number of Distractor Memories} & \multirow{2}{*}{\textbf{PA}} & \multicolumn{2}{c}{\textbf{PF}} & \multirow{2}{*}{\textbf{VRA}} & \multirow{2}{*}{\textbf{CF}} & \multicolumn{2}{c}{\textbf{EI}} \\ \cmidrule(lr){4-8} \cmidrule(lr){10-11} \cmidrule(lr){14-15}
 &  &  & 100 & 300 & 500 & 700 & 1000 &  & Gen. & Int. &  &  & EN & CN \\ \midrule
GPT-5.2 & 0.6720 & \textbf{0.7793} & 0.9519 & 0.9400 & 0.9249 & 0.9138 & 0.8945 & 0.5684 & 0.7043 & \textbf{0.7680} & 0.6640 & 0.9083 & 3.42 & 3.90 \\
GPT-4.1-mini & 0.5373 & 0.6214 & 0.9110 & 0.8578 & 0.8134 & 0.8010 & 0.7700 & 0.4112 & 0.5012 & 0.5240 & 0.2600 & 0.7896 & 2.79 & 2.92 \\
DeepSeek-v3.2 & \textbf{0.6742} & 0.7543 & \textbf{0.9728} & \textbf{0.9687} & \textbf{0.9606} & 0.9333 & \textbf{0.9495} & 0.5912 & 0.6499 & 0.6120 & 0.3540 & \textbf{0.9419} & 3.66 & \textbf{4.00} \\
Gemini-3 Flash & 0.6487 & 0.7686 & 0.9642 & 0.9585 & 0.9484 & \textbf{0.9427} & 0.9440 & \textbf{0.6889} & \textbf{0.7524} & 0.7052 & 0.3000 & 0.8328 & 3.49 & 3.58 \\
Llama-4 Maverick & 0.3772 & 0.6824 & 0.8958 & 0.6290 & 0.5849 & 0.5531 & 0.5275 & 0.2789 & 0.1820 & 0.3080 & 0.7340 & 0.8506 & 2.48 & 2.38 \\
Claude-Sonnet-4.5 & 0.6541 & 0.7543 & 0.9691 & 0.9258 & 0.9437 & 0.9184 & 0.9048 & 0.5649 & 0.5828 & 0.5498 & \textbf{0.8360} & 0.9271 & 3.10 & 3.05 \\
Qwen3-max & 0.6570 & 0.7190 & 0.9402 & 0.8986 & 0.8671 & 0.8216 & 0.7943 & 0.6246 & 0.6981 & 0.6574 & 0.4060 & 0.8267 & \textbf{3.68} & 3.84 \\ \bottomrule
\end{tabular}%
}

\vspace{1mm}

{\scriptsize\parbox{0.95\textwidth}{\textbf{Task 1/2}: semantic-matching F1. \textbf{Task 3}: recall under each distractor-memory setting. \textbf{Task 4}: binary scores for PA/PF/VRA/CF and 1--5 scores for EI. \textbf{PA} = Persona Awareness, \textbf{PF} = Preference Following, \textbf{VRA} = Virtual-Reality Awareness, \textbf{CF} = Constraint Following, \textbf{EI} = Emotional Intelligence.}}
\end{table*}

\subsection{Performance on AlpsBench}
\subsubsection{Task 1: Extraction Result Analysis}
\begin{figure}[t]
\vspace{-0.2cm}
\setlength{\abovecaptionskip}{-0cm}
\setlength{\belowcaptionskip}{-0.3cm}
\centering
\includegraphics[width=\columnwidth]{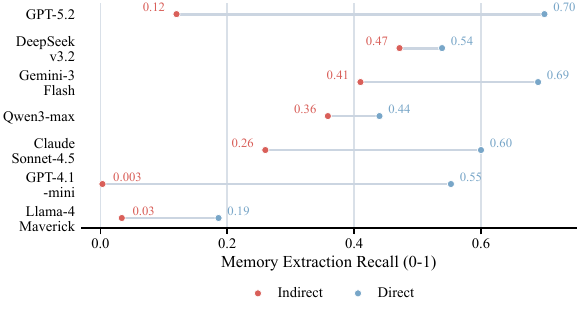}
\caption{Task 1. Direct and indirect memory extraction recall across evaluated models.} 
\Description{Bar chart showing that direct-memory recall is generally higher than indirect-memory recall across the evaluated models.}
\label{fig:direct_indirect}
\end{figure}

\textbf{General-purpose LLMs remain limited in personalized information extraction.}
As shown in Table~\ref{tab:evaluation-results}, the best Task 1 F1 score is achieved by DeepSeek-v3.2 (0.6742), closely followed by GPT-5.2 (0.6720), while Llama-4 Maverick remains substantially lower (0.3772). As shown in Figure~\ref{fig:direct_indirect}, a precision--recall decomposition further indicates that extraction performance is not governed by a single capability. Qwen3-max obtains the highest precision (0.8104) but a relatively lower recall (0.6190), suggesting a conservative extraction behavior; in contrast, Claude-Sonnet-4.5 and Gemini-3 Flash achieve much higher recall (0.8373 and 0.8175) with lower precision. These results show that current LLMs face a persistent trade-off between extracting broader personalized memory sets and maintaining extraction precision.

The direct--indirect split provides a more diagnostic view of this limitation. As shown in Figure~\ref{fig:direct_indirect}, direct memories consistently achieve higher recall than indirect memories across all evaluated models. The gap is relatively small for stronger reasoning models such as DeepSeek-v3.2 (0.5386 vs. 0.4716) and Qwen3-max (0.4399 vs. 0.3586), but becomes severe for comparatively weaker reasoning models: GPT-4.1-mini drops from 0.5526 direct recall to only 0.0032 indirect recall, and Llama-4 Maverick drops from 0.1862 to 0.0337. This pattern suggests that explicit user information can often be captured through surface-level matching, whereas implicit personalized information requires stronger inference over conversational context.

\textbf{Memory-oriented systems improve coverage but still face precision--recall trade-offs.}
As shown in Table~\ref{tab:task1_memsys_exp}, most memory-oriented systems substantially improve recall over the GPT-4.1-mini backbone, suggesting that persistent memory pipelines are effective at preserving a broader set of user-related information. However, this coverage often comes at the cost of precision. EverMemOS achieves the highest recall (0.9340), but its precision remains only 0.0500, leading to a low F1 score of 0.0900. A-Mem and LightMem exhibit the same trade-off: they achieve recall above 0.80, but their low precision leads to limited F1, indicating that higher coverage is obtained at the cost of many false positives. In contrast, Mem0 provides the best overall balance, achieving the highest precision (0.4460) and F1 (0.4570), followed by Mem0g (0.4460 F1) and MemOS (0.4330 F1). These results indicate that simply expanding memory coverage is insufficient for personalized information extraction; effective memory systems must also filter and consolidate stored information to avoid introducing noisy or weakly personalized memories.

\begin{table}[t]
\centering
\caption{Task 1. Extracted information matching results of memory-oriented systems.}
\label{tab:task1_memsys_exp}
\scriptsize
\setlength{\tabcolsep}{4pt}
\renewcommand\arraystretch{1.12}

\resizebox{\columnwidth}{!}{%
\begin{tabular}{lcccccccc}
\toprule
\textbf{ } &
\thead[c]{\textbf{GPT-4.1}\\\textbf{-mini}} &
\thead[c]{\textbf{Ever-}\\\textbf{MemOS}} &
\thead[c]{\textbf{A-Mem}\\\vphantom{g}} &
\thead[c]{\textbf{MemOS}\\\textbf{no-dial}} &
\thead[c]{\textbf{Mem0}\\\vphantom{g}} &
\thead[c]{\textbf{Mem0g}\\\vphantom{g}} &
\thead[c]{\textbf{MemOS}\\\vphantom{g}} &
\thead[c]{\textbf{LightMem}\\\vphantom{g}} \\
\midrule
Precision & 0.3044 & 0.0500 & 0.1370 & 0.2850 & \underline{\textbf{0.4460}} & \underline{0.4010} & \underline{0.3260} & 0.1200 \\
Recall   & 0.3581 & \underline{\textbf{0.9340}} & \underline{0.8830} & \underline{0.8500} & \underline{0.6920} & \underline{0.7230} & \underline{0.8240} & \underline{0.8120} \\
F1       & 0.2970 & 0.0900 & 0.2250 & \underline{0.3890} & \underline{\textbf{0.4570}} & \underline{0.4460} & \underline{0.4330} & 0.1850 \\
\bottomrule
\end{tabular}
}

\vspace{0.5mm}

{\scriptsize\parbox{\columnwidth}{\textbf{no-dial}: MemOS without dialogue context. \textbf{Bold}: best. \underline{Underline}: above GPT-4.1-mini.}}
\end{table}

\subsubsection{Task 2: Update Result Analysis}\label{sec:Task2_exp_result}
\textbf{General-purpose LLMs show a narrower but still visible performance gap on personalized information updating.}
As shown in Table~\ref{tab:evaluation-results}, GPT-5.2 achieves the best Task 2 F1 (0.7793), followed closely by Gemini-3 Flash (0.7686), DeepSeek-v3.2 (0.7543), and Claude-Sonnet-4.5 (0.7543). Compared with Task 1, the top-performing models are more tightly clustered, suggesting that the update task remains challenging but is less sharply differentiating among frontier LLMs under the current evaluation setting. Based on these results, Task 2 suggests that memory updating remains an unresolved challenge: even the strongest models achieve only moderate F1, and the tightly clustered top scores indicate that integrating evolving user information into persistent memory is difficult across frontier LLMs.

\subsubsection{Task 3: Retrieval Result Analysis}

\begin{figure}[t]
\vspace{-0.2cm}
\setlength{\abovecaptionskip}{-0cm}
\setlength{\belowcaptionskip}{0.02cm}
    \centering
    \includegraphics[width=0.72\columnwidth]{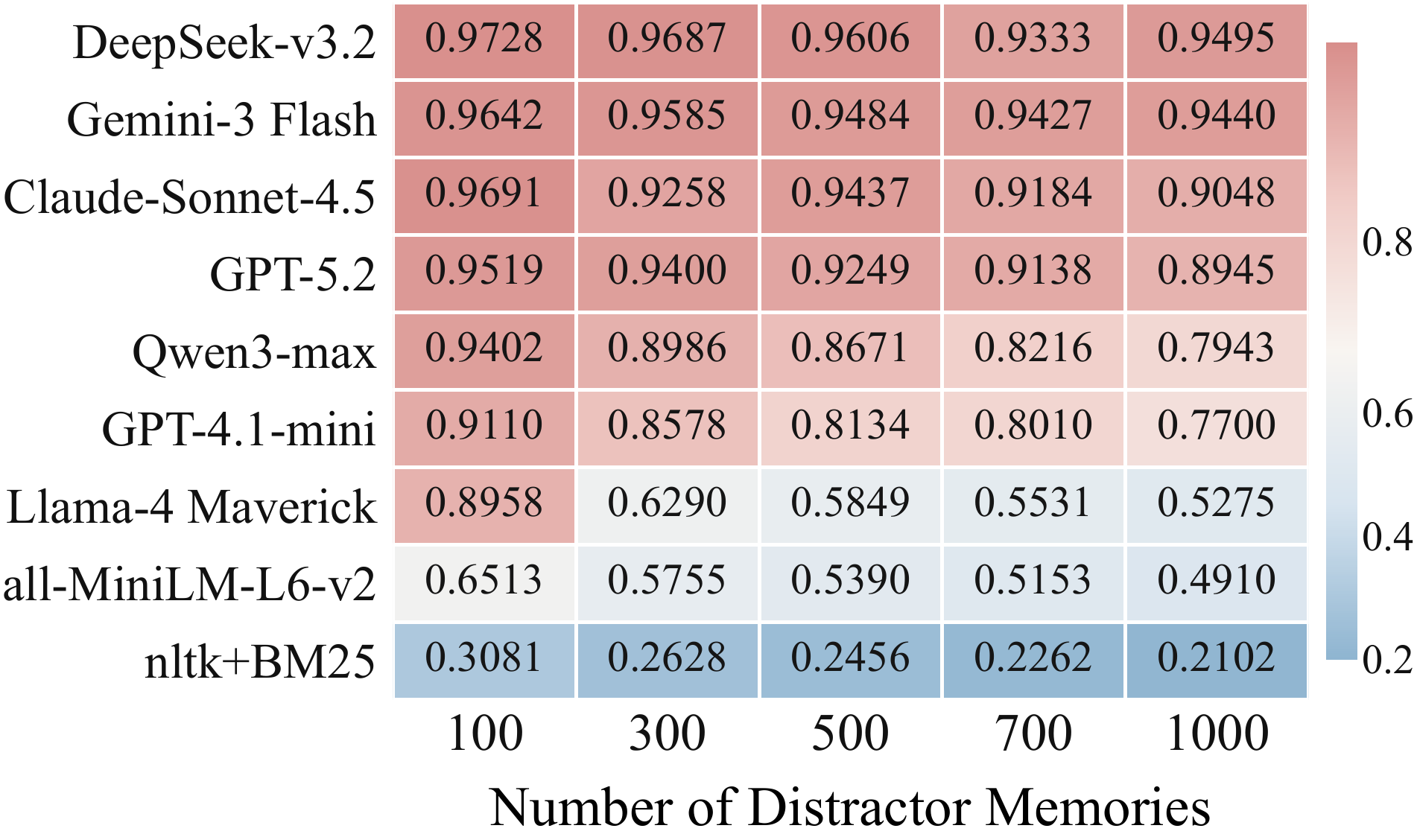}\par
    \makebox[\columnwidth][c]{\scriptsize Models are ordered by their average retrieval score across distractor settings.}
    \caption{Task 3. Heatmap of retrieval scores across evaluated models as the number of distractor memories increases.}
\Description{Heatmap showing Task 3 retrieval recall for each evaluated model under 100, 300, 500, 700, and 1000 distractor memories. Models are ordered by their average retrieval score across distractor settings.}
\label{fig:task3_retrieval_decay}
\end{figure}

\textbf{The sensitivity of general-purpose LLMs to distractor memories varies significantly, with the performance gap between models expanding as distractor density increases.} 
As evidenced by \textbf{Figure~\ref{fig:task3_retrieval_decay}}, all seven evaluated LLMs maintain high and relatively uniform retrieval performance in low-noise environments ($100$ distractors), with scores tightly clustered between $0.8958$ and $0.9728$.
In high-noise settings ($1000$ distractors), the strongest evaluated LLMs remain robust: Gemini-3 Flash and DeepSeek-v3.2 both achieve scores above $0.94$, demonstrating strong resilience in retrieval tasks.
However, given the inherent limitations of context window size and the escalating token costs associated with long-context processing, improving performance in realistic, long-horizon multi-turn scenarios requires a more lightweight and efficient retrieval architecture.

\textbf{Current semantic-level retrieval methods are largely inadequate for personalized memory.}
There is a stark performance disparity between frontier LLMs and the retrieval components typically integrated into memory frameworks. Classical statistical methods such as \textit{nltk+BM25} and semantic embedding baselines like \textit{all-MiniLM-L6-v2} degrade substantially as distractor density scales, with retrieval performance dropping to $0.2102$ and $0.4910$, respectively. This suggests that \quoted{plug-and-play} retrieval modules are fundamentally inadequate for the complexities of long-term personalization. Future research should prioritize the development of logic-aware retrieval layers capable of bringing LLM-level reasoning to the memory retrieval process without the prohibitive costs of full-context processing.

\subsubsection{Task 4: Utilization Result Analysis}
\begin{table}[t]
\setlength{\abovecaptionskip}{0.0cm}
\setlength{\belowcaptionskip}{0.2cm}
\caption{Task 4. Utilization performance comparison of memory-oriented systems.}
\setlength{\tabcolsep}{3.4mm}{
\resizebox{0.48\textwidth}{!}{
\begin{tabular}{llllllll}
\toprule
Method & PA & \multicolumn{2}{c}{PF} & VRA & CF & \multicolumn{2}{c}{EI} \\
\cmidrule(lr){3-4} \cmidrule(lr){7-8}
& & Gen. & Int. & & & EN & CN \\
\midrule
GPT-4.1-mini & 0.4112 & 0.5012 & 0.5240 & 0.2600 & 0.7896 & 2.79 & 2.87 \\
Grd. Mem. & - & - & - & - & - & 3.21 & 3.33 \\
\hdashline \noalign{\smallskip}
MemoryOS & 0.3895 & \underline{0.5195} & \underline{0.5458} & 0.1460 & 0.7872 & 2.23 & 2.26 \\
A-Mem & 0.2895 & \underline{0.5422} & 0.4303 & \textbf{0.2240} & \underline{\textbf{0.8497}} & 1.98 & 1.95 \\
EverMemOS & \underline{\textbf{0.7246}} & \underline{\textbf{0.7888}} & \underline{\textbf{0.6494}} & 0.1667 & 0.7265 & \textbf{2.68} & \textbf{2.73} \\
Mem0 & 0.2561 & 0.4123 & 0.3825 & 0.1640 & \underline{0.7988} & 1.88 & 2.02 \\
Mem0g & 0.2684 & 0.4091 & 0.3904 & 0.1960 & \underline{0.7927} & 1.87 & 1.91 \\
MemOS & 0.2596 & 0.4943 & 0.4502 & 0.1440 & \underline{0.8354} & 2.05 & 2.08 \\
LightMem & 0.2527 & 0.4411 & 0.4025 & 0.1579 & \underline{0.8113} & 1.79 & 1.83 \\
\bottomrule
\end{tabular}
}}
\begin{tablenotes}[flushleft]\scriptsize
\item \textbf{Grd. Mem.} denotes presenting the query together with the ground-truth memories to the LLM.
\end{tablenotes}
\label{tab:performance}
\end{table}


\textbf{Existing LLMs struggle to perform well across all dimensions of personalized memory utilization.}
From the results shown in Table~\ref{tab:evaluation-results}, no single general-purpose LLM consistently dominates all utilization dimensions. Gemini-3 Flash performs best on Persona Awareness (0.6889) and general Preference Following (0.7524), while GPT-5.2 leads on interactive Preference Following (0.7680). Claude-Sonnet-4.5 achieves the strongest Virtual--Reality Awareness (0.8360), DeepSeek-v3.2 performs best on Constraint Following (0.9419) and Chinese Emotional Intelligence (4.00), and Qwen3-max leads English Emotional Intelligence (3.68). This fragmented pattern suggests that personalized agentic competence is multidimensional: models that are strong at aligning with user preferences are not necessarily the most reliable at distinguishing real from virtual memories, following constraints, or producing emotionally appropriate responses.

\textbf{While memory systems enhance model capabilities, they also introduce \quoted{personalization bias.}}
From the results shown in Table~\ref{tab:performance}, memory systems provide clear gains on some utilization dimensions, but the gains are selective. EverMemOS substantially improves Persona Awareness over the GPT-4.1-mini backbone (0.7246 vs. 0.4112) and also improves both general and interactive Preference Following (0.7888 vs. 0.5012; 0.6494 vs. 0.5240). However, these improvements do not transfer uniformly. All evaluated memory systems underperform the backbone on Virtual--Reality Awareness, with the best memory-system score reaching only 0.2240 compared with the backbone's 0.2600. Emotional Intelligence also consistently drops: even the best memory-augmented system reaches only 2.68 in English and 2.73 in Chinese, below the backbone's 2.79 and 2.87. Constraint Following shows a more mixed pattern, where A-Mem performs best (0.8497) but EverMemOS falls below the backbone (0.7265 vs. 0.7896).

These results suggest that current memory systems can make models more persona-aware and preference-sensitive, but may also introduce a form of personalization bias: once retrieved memories are injected, models can over-rely on them, confuse hypothetical or virtual information with real user attributes, and produce less emotionally adaptive responses. Future memory systems should therefore move beyond increasing memory capacity alone and place more emphasis on memory filtering, reliability verification, context grounding, and response-level control, so that personalization improves user alignment without sacrificing factuality, reality awareness, or emotional quality.

\section{Related Work} 

\paragraph{\textbf{LLM Personalization}}
LLM personalization research~\cite{wang2025think,pers2,lin2026bringing,shi2024large,zhao2025nextquill} primarily unfolds along three dimensions~\cite{liu2025survey,pers-survey-1,pers-survey-2,pers-survey-3,pers4,pers5,pers6}. 
Personalized Prompting retrieves user-specific context without modifying parameters, evolving from simple storage to agentic architectures: Mem0~\cite{chhikara2025mem0} and MemInsight~\cite{salama2025meminsight} establish persistent memory layers; A-MEM~\cite{xu2025mem} and Nemori~\cite{nan2025nemori} introduce autonomous decision-making; LightMem~\cite{fang2025lightmem} employs a cognitive \quoted{Sensory-Short-Long} architecture with sleep-time updates; while MemOS~\cite{li2025memos} and MemoryOS~\cite{kang-2025-memory} elevate memory to system-level resources, and EverMemOS~\cite{hu2026evermemos} utilizes Engram-inspired lifecycle management. Personalized Adaptation focuses on parameter-efficient fine-tuning to internalize user patterns. Moving beyond PLoRA~\cite{zhang2024personalized}'s specific adapters, RecLoRA~\cite{zhu2024lifelong} and iLoRA~\cite{kong2024customizing} leverage Mixture-of-Experts with dynamic routing for shifting preferences, while OPPU~\cite{tan2024democratizing} balances privacy via collaborative training. Finally, Personalized Alignment optimizes training objectives: MORLHF~\cite{wu2023fine} improves RL via multi-objective rewards, MODPO~\cite{zhou2024beyond} integrates alignment into DPO, and Personalized Soups~\cite{jang2023personalized} merges parameters at inference for diverse preferences. 

\paragraph{\textbf{LLM Personalization Benchmarks}}
Existing benchmarks can be broadly categorized into two groups: Memory-free and Memory-aware. The first category, represented by LaMP~\cite{salemi2024lamp}, LongLaMP~\cite{kumar2024longlamp}, and LaMP-QA~\cite{salemi2025lamp}, focuses on aligning generated content with user preferences in downstream tasks via retrieved static history. Similarly, PersonalLLM~\cite{zollopersonalllm}, PersonaFeedback~\cite{tao2025personafeedback}, and PersoBench~\cite{afzoon2024persobench} evaluate adherence to pre-defined profiles. However, by treating user history as fixed context, these methods overlook the critical process of personalized information governance (\eg memory extraction, retrieval, and update). In contrast, Memory-aware benchmarks incorporate multi-dimensional assessments of memory capabilities. Early works like EmoBench~\cite{emobench} and EQ-Bench~\cite{eqbench} targeted emotional alignment, while LoCoMo~\cite{locomo} and LongMemEval~\cite{longmemeval} target long-term memorization. Specific diagnostics such as PrefEval~\cite{zhaollms} and HaluMem~\cite{chen2025halumem} further address preference consistency and memory hallucinations. Recently, agentic simulations like PersonaLens~\cite{zhao2025personalens}, PersonaMem~\cite{jiang2025know}, and its successor PersonaMem v2~\cite{jiang2025personamem} have been proposed to track profile evolution. Despite these advances, their reliance on synthetic human-LLM dialogues creates a significant distributional gap from real-world data, resulting in homogeneous conversations~\cite{jiang2025artificial} that lack natural diversity and fail to capture the implicit expressions characteristic of authentic human interactions.

\section{Conclusion and Future Work}
In this paper, we presented {AlpsBench}, a benchmark designed to evaluate the complete lifecycle of LLM personalization using real-world dialogue data. By leveraging long-term human-LLM interactions and expert-verified structured memories, AlpsBench enables a granular diagnosis of LLM performance across four key dimensions: personalized information extraction, update, retrieval, and utilization. Extensive experiments highlight several critical challenges for current LLMs, including difficulties in interpreting implicit user information, handling preference drift, and maintaining retrieval reliability under heavy interference.
Going forward, we will continuously maintain and update the dataset, making it a dynamic benchmark to evaluate emerging LLMs. 

\begin{acks}
This work was supported by the National Natural Science Foundation of China (U25B2071 and 62525211).
\end{acks}

\bibliographystyle{ACM-Reference-Format}
\bibliography{references}




\end{document}